%% file: neurips_2026.tex
\documentclass{article}

\PassOptionsToPackage{numbers, compress}{natbib}

\usepackage[main, final]{neurips_2026}

\usepackage[utf8]{inputenc} 
\usepackage[T1]{fontenc}    
\usepackage{hyperref}       
\usepackage{url}            
\usepackage{booktabs}       
\usepackage{amsfonts}       
\usepackage{nicefrac}       
\usepackage{microtype}      
\usepackage{xcolor}         
\definecolor{my_green}{rgb}{0.278,0.408,0.376}
\definecolor{anthropic}{rgb}{0.800,0.486,0.365}
\definecolor{cityu}{rgb}{0.745,0.184,0.251}
\definecolor{my_red}{rgb}{0.592,0.286,0.274}
\hypersetup{bookmarks=true,colorlinks=true,linkcolor=cityu,urlcolor=cityu,citecolor=cityu}

\usepackage{url}
\usepackage{booktabs}       
\usepackage{amsfonts}       
\usepackage{nicefrac}       
\usepackage{microtype}      
\usepackage{amsmath}        
\usepackage{multirow}
\usepackage{bigdelim}
\usepackage{makecell}
\usepackage{listings}
\usepackage{upquote}
\usepackage[normalem]{ulem} 
\usepackage{subcaption}
\usepackage{pifont}
\usepackage{colortbl}
\usepackage{enumitem}

\usepackage[ruled, vlined, linesnumbered]{algorithm2e}
\SetNlSkip{1em} 
\usepackage{wrapfig}

\usepackage{tcolorbox}
\tcbuselibrary{listings, breakable} 
\usepackage{listings}

\title{Tree-Based Sampling for Reinforcement Learning in LLM-Driven Automated Algorithm Design}
\title{Tree Sampling for Reinforcement Learning in Automated Algorithm Design with LLMs}

\author{%
  Rui Zhang$^{1}$,\quad
  Ping Guo$^{2}$\thanks{Corresponding author.},\quad
  Liyong Lin$^{3}$,\quad
  Zhichao Lu$^{1}$,\quad
  Qingfu Zhang$^{1}$
  \\
  $^{1}$Department of Computer Science, City University of Hong Kong \\
  $^{2}$Hong Kong Yuwin Technology Co., Ltd \\
  $^{3}$Contemporary Amperex Technology Limited \\
  \fontsize{10.2pt}{0.1\baselineskip}\selectfont
  \texttt{rui.zhang.cs@my.cityu.edu.hk,}\quad
  \texttt{pingguo@yuyi-hk.com,}\\
  \texttt{llin5@e.ntu.edu.sg,}\quad
  \texttt{\{zhichao.lu, qingfu.zhang\}@cityu.edu.hk}
}

\begin{document}

\maketitle

\input{src/0-abstract}
\input{src/1-introduction}
\input{src/2-background}
\input{src/3-method}
\input{src/4-experiment}
\input{src/5-conclusion}
\begin{ack}
This work received no specific funding. 
The authors declare no competing interests.
\end{ack}

\bibliographystyle{plainnat}
\bibliography{mybib}


\appendix
\input{src/6-appendix}


\newpage
\input{checklist.tex}

\end{document}

%% file: src/0-abstract.tex
\begin{abstract}
Large Language Model-based Automated Algorithm Design (LLM-AAD) has shown promising results across diverse domains by coupling LLMs with iterative search. 
Recent efforts have begun to combine this paradigm with reinforcement learning (RL),
using reward feedback from algorithm evaluators to update the LLM and better guide subsequent algorithm generation. 
However, these methods still rely primarily on outcome-level supervision. As a result, intermediate ideas and plans receive no direct credit, even though they guide the direction of algorithm design. 
We introduce \emph{\uline{\textbf{A}}lgorithm \uline{\textbf{T}}ree \uline{\textbf{P}}olicy \uline{\textbf{O}}ptimization (ATPO)}, a reinforcement learning method that can be integrated into the algorithm-generation stage of existing LLM-based automated algorithm design methods.
ATPO employs tree-based sampling to expand each idea into multiple executable code implementations, using their evaluation outcomes to estimate idea-level quality and provide structured credit at both the idea and code levels.
Experiments integrating ATPO with FunSearch, EoH, and OpenEvolve demonstrate improvements over their original counterparts.
The learned policy also transfers to held-out problem instances and alternative search methods, suggesting its potential to generalize within the same AAD task.
Source code is available at \url{https://github.com/RayZhhh/atpo-aad}.
\end{abstract}

%% file: src/1-introduction.tex
\section{Introduction}
The emerging field of Large Language Model-based Automated Algorithm Design (LLM-AAD) has attracted growing attention for its potential to generate expert-level algorithms~\cite{liu2024systematic,romera2024mathematical}.
Within this field, a prevailing paradigm couples LLMs with iterative search: the LLM serves as a generator that proposes candidate algorithms, while the search procedure evaluates, selects, and feeds them back to guide the next round of generation~\cite{liu2024evolution,romera2024mathematical, novikov2025alphaevolve}.
This paradigm has led to notable advances across a spectrum of algorithmic development tasks, 
including machine learning~\cite{ma2024eureka, lu2024discovering, yang2025r, du2026mlevolve}, combinatorial optimization~\cite{liu2024evolution, ye2024reevo}, and scientific discovery~\cite{romera2024mathematical, ma2024llm, shojaee2025llmsr, novikov2025alphaevolve}.

\begin{figure}[t]
    \centering
    \includegraphics[width=\linewidth]{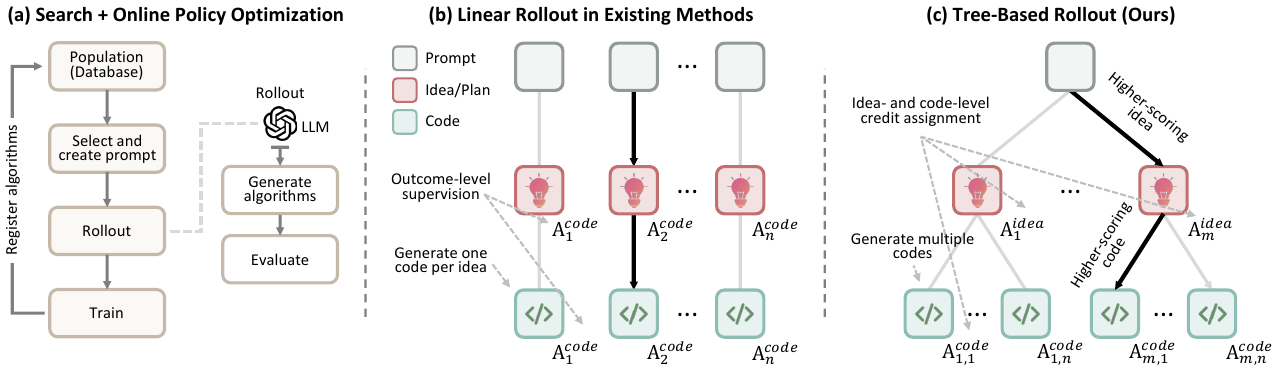}
    \caption{
    Overview of the LLM-AAD framework with online policy optimization and comparison of rollout structures.
    \textbf{(a)} The search framework repeatedly selects existing algorithms to construct prompts that guide the generation of new algorithms, while evaluator feedback from the online rollout is used to optimize the policy.
    \textbf{(b)} Existing methods follow a linear rollout, with each sampled trajectory containing one idea/plan and one code candidate. The training signal is derived solely from the evaluation of the final code candidate.
    \textbf{(c)} Our tree-based rollout branches each prompt into multiple ideas and each idea into multiple code implementations, enabling idea-level potential estimation from multiple implementations and separate training signals for ideas and code.
    }
    \label{fig:rollout_comparison}
\end{figure}

Despite these advances, most existing LLM-AAD approaches rely on off-the-shelf LLMs, which pose two key limitations:
\ding{182}~improvements come entirely from better prompts and better outer-loop search, while the generator itself remains untouched; and
\ding{183}~these LLMs are not trained to generate reusable algorithmic ideas or to improve from evaluator feedback collected online.
Recent efforts like EvoTune~\cite{surina2025algorithm}, CALM~\cite{huang2025calm}, and ThetaEvolve~\cite{wang2025thetaevolve} extend algorithm search with online policy optimization, using raw trajectories generated during the search to provide evaluator feedback to improve the LLM.
Figure~\ref{fig:rollout_comparison}(a) provides a conceptual overview of this paradigm: the search process continually generates and evaluates candidate algorithms, while the resulting rollout data is used to update the policy and guide subsequent generations.
However, these attempts supervise the generator only at the \emph{outcome-level}: the reward depends solely on the algorithm's performance, 
while the intermediate reasoning traces, such as ideas and proposed solution plans, receive no direct credit.
More critically, ideas and plans occupy a central role in algorithm design: they determine the conceptual direction of search and govern which regions of the solution space are explored~\cite{wang2024planning}. 
Consequently, the explicit rewarding of ideas and plans is crucial, but its reliable estimation demands a specialized rollout structure that existing methods do not support.

As illustrated in Figure~\ref{fig:rollout_comparison}(b), existing methods adopt a \emph{linear} rollout: each sampled idea is expanded into a single code implementation.
However, algorithm design inherently demands broad exploration, as a single conceptual idea can be instantiated in vastly different ways through different implementations and hyperparameter settings. 
In a linear rollout, a single code implementation yields an unreliable proxy for the underlying idea's true potential.
For example, a promising idea may receive a severely degraded reward due to a trivial coding error or an unfortunate hyperparameter choice.
To address this, 
we use a tree-structured rollout that expands each idea into multiple executable code candidates.
As illustrated Figure~\ref{fig:rollout_comparison}(c), the tree-structured rollout expands each idea into multiple executable code candidates, so that its achievable implementation potential can be estimated more reliably from the performance of multiple code implementations. 

In this paper, we propose \emph{\uline{\textbf{A}}lgorithm \uline{\textbf{T}}ree \uline{\textbf{P}}olicy \uline{\textbf{O}}ptimization (ATPO)}, a reinforcement learning method that can be integrated into the algorithm generation stage of existing LLM-AAD search frameworks. 
As illustrated in Figure~\ref{fig:algo_tree_overview}, once a prompt has been constructed by the search framework, ATPO performs tree-based, two-level exploration by sampling multiple ideas under each prompt and multiple code implementations under each idea.
Code leaves are evaluated by the task evaluator and scored, and then aggregated upward to estimate an idea-level potential score.
ATPO performs a two-level advantage estimation: 
At the idea level, advantages are computed \emph{across idea segments} under the same prompt, rewarding conceptual directions that consistently yield better algorithms.
At the code level, advantages are computed independently \emph{across code segments} under the same idea, rewarding implementations that effectively instantiate it.
Empirically, integrating ATPO with three representative search frameworks can improve performance relative to their original counterparts, and the learned policy generalizes to unseen problem instances and alternative search methods.
Our main contributions are summarized as follows:
\begin{itemize}[leftmargin=*]
    \item We propose ATPO, a reinforcement learning method for automated algorithm design that performs two-level advantage estimation across ideas generated under the same prompt and across code implementations of the same idea. These two signals jointly guide policy optimization at the idea and code levels using feedback from code evaluation.
    \item ATPO can be integrated into existing LLM-AAD search frameworks.
    In this work, we instantiate it with three representative methods: FunSearch~\cite{romera2024mathematical}, EoH~\cite{liu2024evolution}, and AlphaEvolve~\cite{novikov2025alphaevolve}.
    The policy learns from evaluator feedback collected during search, and the updated policy generates candidate programs for subsequent search iterations.
    \item We evaluate ATPO on three algorithm-design benchmarks (Admissible Set Problem, Online Bin Packing, and Traveling Salesman Problem) and demonstrate improvements over their original counterparts. 
    We further demonstrate the learned policy’s transfer to held-out problem instances and alternative search methods within the same task.
\end{itemize}

%% file: src/2-background.tex
\section{Preliminary}
\label{sec:prelim}

\subsection{LLM-Based Automated Algorithm Design}

LLM-based automated algorithm design (LLM-AAD) aims to synthesize a high-performing algorithm implementation for a target task.
Concretely, the goal is to produce a program $f$ that maximizes the expected reward $\mathbb{E}_{x \sim \mathcal{D}}[r(f, x)]$ over a distribution $\mathcal{D}$ of problem instances, where $r$ is a task-specific reward function that assigns a scalar score to program $f$ on instance $x$.

\paragraph{Search Framework.}
Methods in this area integrate an LLM into an iterative
search framework~\cite{romera2024mathematical,liu2024evolution, ye2024reevo, yao2025multi, zheng2025monte, dat2025hsevo, lange2025shinkaevolve, zhang2026rethinking}.
As illustrated in Figure \ref{fig:algo_tree_overview}(a), the search method maintains a population (or archive) $\mathcal{P}$ of discovered programs together with their evaluation scores.
At each iteration, the controller selects a subset of programs from $\mathcal{P}$ using selection strategies and combines them with the task description to form a prompt $p$. 
The LLM generates one or more new candidate programs conditioned on $p$. Each candidate is executed, evaluated by $r$, and registered back into $\mathcal{P}$ according to the search method's update rule.

Representative search frameworks mainly differ in how they organize the population and construct prompts.
FunSearch~\cite{romera2024mathematical} uses a multi-island database to preserve diversity.
EoH~\cite{liu2024evolution} introduces evolutionary operators that generate both high-level ideas and their code implementations.
AlphaEvolve~\cite{novikov2025alphaevolve} further strengthens population management with finer prompt construction and niche-based diversity control.

\subsection{Tree-Based Rollout for Reinforcement Learning}
Recent work has explored tree-based rollouts for reinforcement learning in LLM post-training to address sparse and delayed rewards and improve exploration~\cite{zhang2024rest, hou2025treerl, ji2025tree, guo2025segment, li2025treepo, kazemnejad2024vineppo, feng2023alphazero}.
Rather than treating each sampled completion as an independent chain, these methods organize rollouts into trees so that promising prefixes can be expanded and reused, yielding more informative intermediate supervision. 
Existing approaches instantiate this idea at different granularities, such as reasoning traces~\cite{zhang2024rest, hou2025treerl}, agent steps~\cite{ji2025tree}, or response segments~\cite{guo2025segment, li2025treepo}. 
This line of work suggests that the key choice is not only whether to branch, but also which unit should receive credit.

ATPO adapts this perspective to LLM-AAD, where the natural semantic units are the \emph{idea/plan} and the resulting \emph{code implementation}. 
It therefore explores multiple ideas under the same prompt and multiple implementations of the same idea, and assigns separate training signals to these two segments. 
Unlike methods that partition responses into generic steps or fixed-length segments, our segmentation follows the semantic structure of algorithm generation. 
This design estimates each idea's potential from multiple code implementations, providing a more reliable basis for assigning credit to intermediate ideas during online policy optimization within an LLM-AAD search loop.

%% file: src/3-method.tex
\begin{figure}[t]
    \centering
    \includegraphics[width=\linewidth]{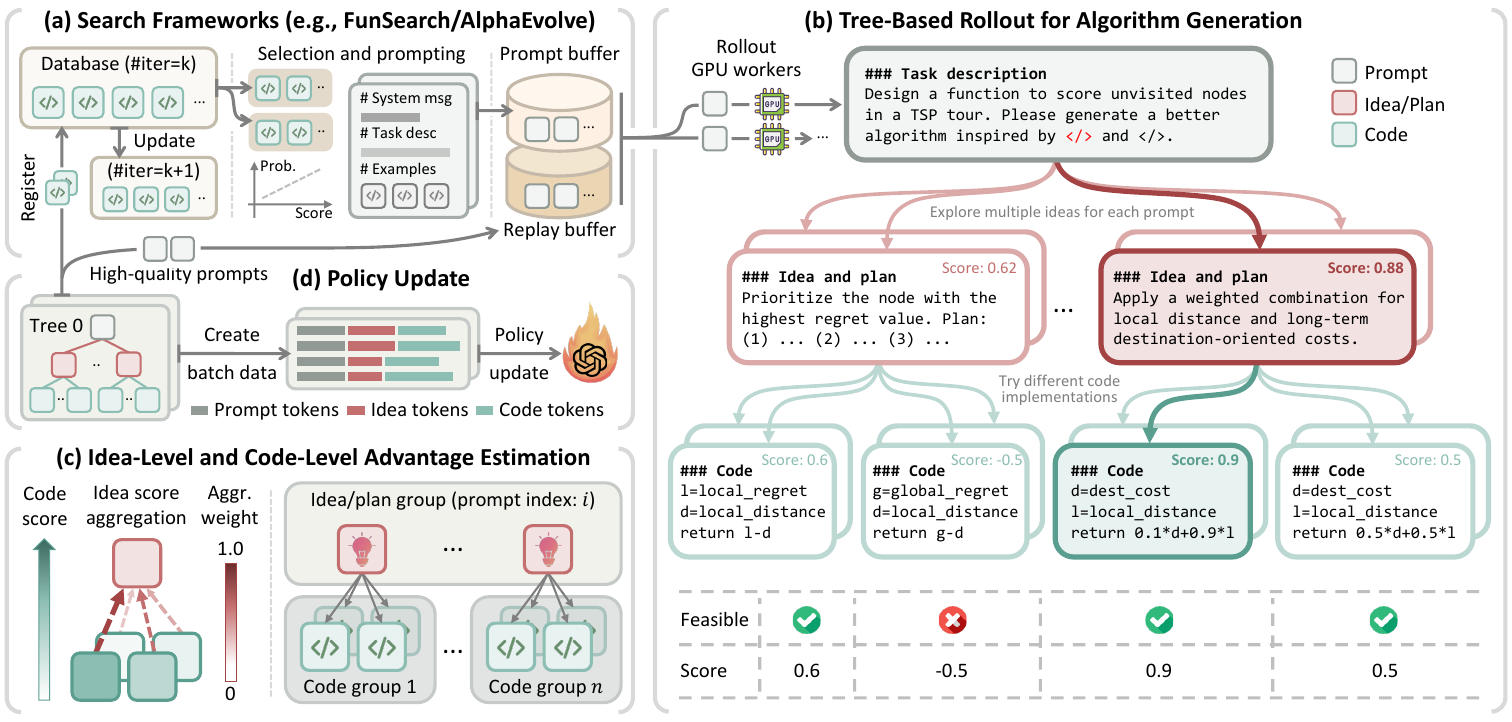}
    \caption{
    Overall pipeline integrating ATPO into an LLM-AAD search framework.
    \textbf{(a)} Existing LLM-AAD search methods, such as FunSearch and AlphaEvolve, maintain an evolving population of candidate programs, iteratively select promising algorithms to construct prompts, request LLMs to generate new algorithms, and update the population for future exploration. 
    \textbf{(b)} For each selected prompt, we perform a tree-structured rollout: we first generate multiple idea candidates and then expand each idea into multiple code implementations, which are executed and scored by the task evaluator.
    \textbf{(c)} Code-level scores are computed for each leaf and aggregated upward into idea-level scores.
    We estimate idea-level advantages by comparing ideas generated under the same prompt, and code-level advantages by comparing implementations of the same idea.
    \textbf{(d)} The resulting prompt, idea, and code tokens are assembled into training batches for policy optimization. Valid algorithms and breakthrough prompts are registered in the population and the replay buffer, respectively.
    }
    \label{fig:algo_tree_overview}
\end{figure}

\section{Method}
\label{sec:method}

In this section, we introduce ATPO, a reinforcement learning method that can be integrated into the algorithm generation stage of existing LLM-AAD search frameworks. 
Figure~\ref{fig:algo_tree_overview} summarizes the full pipeline. 
Built on top of existing search backbones such as FunSearch, EoH, and AlphaEvolve~\cite{romera2024mathematical,liu2024evolution,novikov2025alphaevolve}, ATPO keeps the search loop intact while using the resulting trajectories to optimize the LLM policy.
The remainder of this section covers the algorithm's design, including the tree-structured rollout procedure, the segment-level advantage estimation, and the policy update. Engineering details are provided in Appendix~\ref{app:implementation}.

\subsection{Tree-Based Sampling}
At training step $t$, the controller selects a batch of prompts $\{p_i\}_{i=1}^B$. Each prompt contains the task description, along with the search context inherited from the underlying search method, such as previously discovered programs and their metadata. 
As shown in Figure~\ref{fig:algo_tree_overview}(b), each selected prompt is expanded into a two-level rollout tree rather than a single linear completion.
Given a selected prompt $p_i$, the policy samples $N_I$ idea candidates,
$$
z^{\mathrm{idea}}_{i,1}, \ldots, z^{\mathrm{idea}}_{i,N_I},
$$
where each idea is a natural-language plan for improving the target algorithm. Conditioned on the prompt and one sampled idea, the policy then samples $N_C$ code candidates,
$$
z^{\mathrm{code}}_{i,j,1}, \ldots, z^{\mathrm{code}}_{i,j,N_C},
$$
which instantiate idea $j$ into executable programs. Each leaf trajectory can therefore be written as
$$
\tau_{i,j,k} = \left(p_i, z^{\mathrm{idea}}_{i,j}, z^{\mathrm{code}}_{i,j,k}\right).
$$
Concretely, each leaf trajectory is associated with two grouping identifiers: a prompt identifier that links all leaves generated from the same prompt, and an idea identifier that links all code candidates derived from the same idea. 
These identifiers are used later for advantage estimation.
In practice, the tree is built in two phases: each idea and an initial code candidate are generated jointly in a single forward pass. The remaining $N_C - 1$ code candidates per idea are then generated by conditioning on the same prompt--idea context. 

The rollout is robust to invalid generations. 
If a generated sequence does not conform to the prescribed output format, it cannot be parsed into valid idea and code nodes and is treated as an invalid sequence.
Such sequences are still kept in the batch because they provide direct supervision of the current policy's failure modes.
Due to the space limit, readers are referred to Appendix~\ref{app:tree-gen} for more details.

\subsection{Code-Level Evaluation and Idea-Level Aggregation}
For each leaf node (code node), we extract the program within it and submit it to the task evaluator to obtain its score. The raw score $s_{i,j,k}$ is assigned as
$$
s_{i,j,k} =
\begin{cases}
-1   & \text{parse failure (missing format markers or extraction error),} \\
-0.5 & \text{runtime error or non-finite evaluator output,} \\
r_{\mathrm{eval}} & \text{valid execution,}
\end{cases}
$$
where $r_{\mathrm{eval}} \in [0, 1]$ is the task evaluator's output, normalized relative to a reference baseline and a performance bound. The task-specific normalization is described in Appendix~\ref{app:reward}.

As illustrated in Figure~\ref{fig:algo_tree_overview}(c), after evaluating the code leaves, we aggregate their scores upward to the corresponding parent idea nodes, thereby constructing idea-level supervision from code-level outcomes.
In algorithm design, an idea's value lies in its achievable potential when instantiated by executable implementations, not in the average quality of its implementations. 
We therefore weight higher-scoring child codes more heavily when forming the idea-level score, rather than averaging all siblings uniformly.
We then aggregate code-level scores into an idea-level score. For idea node $(i,j)$, let $\mathcal{V}_{i,j}$ be the set of valid child codes, and let $\operatorname{TopK}(\mathcal{V}_{i,j})$ denote the top-$K$ scores in that set. 
The idea score is computed by a temperature-controlled softmax aggregation,
$$
q_{i,j} = \sum_{k \in \operatorname{TopK}(\mathcal{V}_{i,j})}
\frac{\exp(s_{i,j,k}/\tau)}{\sum_{k' \in \operatorname{TopK}(\mathcal{V}_{i,j})}\exp(s_{i,j,k'}/\tau)}
s_{i,j,k}.
$$
This top-$K$ aggregation reduces the influence of weak implementations while still preserving information from multiple strong descendants.

\subsection{Two-Level Credit Assignment with GRPO}
ATPO optimizes the policy at two levels: the idea-level objective encourages generating higher-scoring algorithmic ideas/plans, while the code-level objective encourages exploring higher-reward implementations for each given idea/plan. The two group-relative normalization groups below formalize this two-level credit assignment.
This two-level grouping follows the prompt-level and idea-level organization shown in Figure~\ref{fig:algo_tree_overview}(c).

We therefore compute two separate group-relative advantages. The \emph{idea advantage} compares different ideas generated under the same prompt, while the \emph{code advantage} compares different code implementations generated under the same idea. Let
$$
\mathcal{G}_i = \{\tau_{i,j,k}\}_{j,k}
\qquad\text{and}\qquad
\mathcal{H}_{i,j} = \{\tau_{i,j,k}\}_k
$$
denote the prompt-level and idea-level groups, respectively. Let $R^{\mathrm{idea}}_{i,j,k} = q_{i,j}$ and $R^{\mathrm{code}}_{i,j,k} = s_{i,j,k}$ be the idea-level and code-level returns. We compute
$$
A^{\mathrm{idea}}_{i,j,k} =
\frac{R^{\mathrm{idea}}_{i,j,k} - \mu^{\mathrm{idea}}_i}{\sigma^{\mathrm{idea}}_i + \epsilon},
\qquad
A^{\mathrm{code}}_{i,j,k} =
\frac{R^{\mathrm{code}}_{i,j,k} - \mu^{\mathrm{code}}_{i,j}}{\sigma^{\mathrm{code}}_{i,j} + \epsilon},
$$
where $(\mu^{\mathrm{idea}}_i,\sigma^{\mathrm{idea}}_i)$ are computed over $\mathcal{G}_i$, and $(\mu^{\mathrm{code}}_{i,j},\sigma^{\mathrm{code}}_{i,j})$ are computed over $\mathcal{H}_{i,j}$. 

Let $\pi_\theta$ be the current policy, $\pi_{\theta_{\mathrm{old}}}$ the behavior policy used to collect trajectories, and $y_{i,j,k,t}$ the $t$-th generated token in trajectory $\tau_{i,j,k}$. Following GRPO~\cite{shao2024deepseekmath}, we define the token-wise importance ratio
$$
r_{i,j,k,t}(\theta) =
\frac{\pi_\theta\!\left(y_{i,j,k,t}\mid p_i, y_{i,j,k,<t}\right)}
{\pi_{\theta_{\mathrm{old}}}\!\left(y_{i,j,k,t}\mid p_i, y_{i,j,k,<t}\right)}.
$$

\paragraph{Weighted Broadcast Advantage.}
The first variant collapses the two group-relative signals into a single scalar advantage,
$$
A^{\mathrm{weighted}}_{i,j,k}
=
\alpha A^{\mathrm{idea}}_{i,j,k} + (1-\alpha) A^{\mathrm{code}}_{i,j,k},
$$
where $\alpha \in (0,1)$ balances achievable potential at the idea level against implementation quality at the code level. This mixed scalar is then broadcast to all valid response tokens of trajectory $\tau_{i,j,k}$:
$$
\hat{A}^{\mathrm{weighted}}_{i,j,k,t}
=
A^{\mathrm{weighted}}_{i,j,k},
\qquad t=1,\ldots,T_{i,j,k},
$$
where $T_{i,j,k}$ is the total number of generated tokens in $\tau_{i,j,k}$.
The same weighted combination is applied to the returns. The actor is optimized with the standard PPO-style clipped objective
$$
\begin{aligned}
\mathcal{L}^{\mathrm{weighted}}_{\mathrm{ATPO}}(\theta)
&=
\mathbb{E}_{\tau_{i,j,k}}
\Bigg[
\frac{1}{T_{i,j,k}}
\sum_{t=1}^{T_{i,j,k}}
\min\!\Big(
r_{i,j,k,t}(\theta)\hat{A}^{\mathrm{weighted}}_{i,j,k,t}, \\
&\qquad\qquad
\operatorname{clip}\!\left(r_{i,j,k,t}(\theta), 1-\varepsilon, 1+\varepsilon\right)
\hat{A}^{\mathrm{weighted}}_{i,j,k,t}
\Big)
\Bigg].
\end{aligned}
$$

\paragraph{Segment-Differentiated Advantage.}
The second variant keeps the two signals separate and assigns them to different response segments, following the same segment-wise credit-allocation intuition as SPO. Let $e_{i,j,k}$ denote the boundary between the idea and code segments, i.e., the first code token position in the response. We define
$$
m^{\mathrm{idea}}_{i,j,k,t} = \mathbb{I}[t < e_{i,j,k}],
\qquad
m^{\mathrm{code}}_{i,j,k,t} = \mathbb{I}[t \ge e_{i,j,k}],
$$
so that idea tokens and code tokens are activated by disjoint masks. The token-level training signal is then
$$
\hat{A}^{\mathrm{diff}}_{i,j,k,t}
=
m^{\mathrm{idea}}_{i,j,k,t} A^{\mathrm{idea}}_{i,j,k}
+
m^{\mathrm{code}}_{i,j,k,t} A^{\mathrm{code}}_{i,j,k}.
$$
Thus, idea tokens are optimized only with the inter-idea comparison signal, while code tokens are optimized only with the within-idea comparison signal. In practice, if an idea has only one sampled code implementation, we set $A^{\mathrm{code}}_{i,j,k}=0$ because no within-idea comparison is available. The resulting objective can be written in a segment-wise form analogous to the SPO~\cite{guo2025segment} policy-gradient objective:
$$
\begin{aligned}
\mathcal{L}^{\mathrm{diff}}_{\mathrm{ATPO}}(\theta)
&=
\mathbb{E}_{\tau_{i,j,k}}
\Bigg[
\frac{1}{T_{i,j,k}}
\sum_{s \in \{\mathrm{idea}, \mathrm{code}\}}
\sum_{t=1}^{T_{i,j,k}}
m^s_{i,j,k,t}
\min\!\Big(
r_{i,j,k,t}(\theta) A^s_{i,j,k}, \\
&\qquad\qquad
\operatorname{clip}\!\left(r_{i,j,k,t}(\theta), 1-\varepsilon, 1+\varepsilon\right)
A^s_{i,j,k}
\Big)
\Bigg],
\end{aligned}
$$
where $A^{\mathrm{idea}}_{i,j,k}$ and $A^{\mathrm{code}}_{i,j,k}$ are the idea-level and code-level normalized advantages, respectively. In other words, the loss is decomposed over two semantic segments rather than being driven by a single trajectory-wide scalar.

When a reference policy is used, we optionally add a KL regularizer to either variant, yielding
$$
\mathcal{L}_{\mathrm{actor}}(\theta)
=
\mathcal{L}^{\mathrm{variant}}_{\mathrm{ATPO}}(\theta)
- \beta \,
\mathbb{E}_{\tau_{i,j,k}}
\left[
\frac{1}{T_{i,j,k}}\sum_{t=1}^{T_{i,j,k}}
D_{\mathrm{KL}}\!\left(
\pi_\theta(\cdot \mid p_i, y_{i,j,k,<t})
\;\|\;
\pi_{\mathrm{ref}}(\cdot \mid p_i, y_{i,j,k,<t})
\right)
\right].
$$
This objective directly encourages the policy to generate ideas with higher achievable implementation potential given a prompt, and high-quality implementations given an idea, ultimately maximizing the evaluator reward.
As a result, both the weighted-broadcast and segment-differentiated variants optimize the same two-level training objective, and we compare their empirical performance in the ablation study (Section~\ref{sec:ablation}).

\subsection{Register: Writing Back to Search Framework and Replay}
\label{sec:register}
After each training step, all valid programs are written back into the search method, which applies its own population-management rule. In FunSearch, for instance, each new program is inserted into the island of its parent candidate. 

Separately, we maintain a replay buffer of \emph{breakthrough prompts}, namely prompt contexts that produce a new historical best score. 
As shown in Figure~\ref{fig:algo_tree_overview}(a,d), this buffer complements the main search population by storing useful prompt contexts rather than only good programs. When the search stagnates, the controller can temporarily switch from the standard prompt selector to this buffer, revisiting prompt contexts that are responsible for earlier progress. 
Each replayed prompt has a finite lifetime controlled by a usage budget $U_{\max}$. In all experiments, we set $U_{\max}=1$, so each stored prompt is replayed at most once. Due to the space limit, further implementation details are demonstrated in Appendix~\ref{app:replay-buffer}.

%% file: src/4-experiment.tex
\section{Experiments} \label{sec:experiments}

In this section, we investigate whether ATPO can improve the performance of LLM-AAD along the following two dimensions:
\begin{itemize}[leftmargin=*]
    \item \textbf{RQ1:} Does integrating ATPO into existing LLM-AAD search frameworks improve search performance under a fixed evaluation budget? Specifically, can online policy optimization use evaluator feedback to guide the LLM toward higher-quality algorithms, as reflected by a lower final average relative gap, across different search frameworks?
    \item \textbf{RQ2:} Does the policy learned with ATPO generalize beyond its training setting? We test whether an ATPO-trained model remains effective when combined with a different search method or applied to held-out problem instances, without additional task-specific fine-tuning.
\end{itemize}

\paragraph{AAD Tasks.}
We evaluate ATPO on three widely used AAD benchmarks spanning constrained set construction, online decision making, and constructive routing: Admissible Set Problem (ASP)~\cite{romera2024mathematical}, Online Bin Packing (OBP) \cite{seiden2002online}, and constructive Traveling Salesman Problem (TSP) \cite{matai2010traveling}. 
These evaluators are adapted from the LLM4AD platform~\cite{liu2024llm4ad}.
Due to space limitations, the task descriptions, seed algorithms, and reward formulations are detailed in Appendix~\ref{app:task-prompts}.

\paragraph{Experimental Settings.} 
We integrate ATPO with three representative LLM-AAD methods, FunSearch~\cite{romera2024mathematical}, EoH~\cite{liu2024evolution}, and OpenEvolve~\cite{openevolve} (which is an open-source implementation of AlphaEvolve~\cite{novikov2025alphaevolve}).
We use Qwen2.5-1.5B-Instruct~\cite{qwen2.5, qwen2} and Qwen2.5-7B-Instruct~\cite{qwen2.5, qwen2} as the base models, hereafter abbreviated as Qwen2.5-1.5B and Qwen2.5-7B, respectively.
Each LLM is fine-tuned with LoRA~\cite{hu2022lora}, with rank 32 and an alpha parameter of 16. We perform online RL directly from the instruction-tuned models without a preceding SFT stage or an external SFT dataset. Supervision comes solely from evaluator feedback on generated code.
For ATPO, we set $N_I=N_C=4$ for tree-structured rollout and $K=2$ for idea scoring. The standard GRPO baseline uses a group size of 16, matching the $N_I N_C=16$ code candidates per prompt in ATPO. Both methods use the same total budget of 20,000 evaluated algorithms. This comparison aligns candidate counts and evaluator calls, rather than wall-clock time or total compute. Our ATPO-AAD training framework is implemented based on the \texttt{verl} library~\cite{sheng2024hybridflow}.
The experiments are executed on a GPU server equipped with eight NVIDIA H20 GPUs. Qwen2.5-1.5B is trained on a single GPU, while Qwen2.5-7B is trained on two GPUs for acceleration.

\begin{table}[h]
\caption{Performance comparison of the top-5 algorithms for ASP, OBP, and TSP. Results are reported as mean $\pm$ standard deviation of the average relative gap (\%, lower is better) over three independent runs. The best mean result for each setting is highlighted in \textbf{bold}.}
\centering
\begingroup
\small
\begin{tabular}{lccccc}
\toprule
\multirow{2}{*}{Method} & \multicolumn{2}{c}{ASP} & \multicolumn{2}{c}{OBP} & TSP \\
\cmidrule(lr){2-3}\cmidrule(lr){4-5}\cmidrule(lr){6-6}
 & Qwen2.5-1.5B & Qwen2.5-7B & Qwen2.5-1.5B & Qwen2.5-7B & Qwen2.5-7B \\
\midrule
FunSearch & $7.43_{\pm 0.24}$ & $6.91_{\pm 0.53}$ & $\mathbf{0.57}_{\pm 0.02}$ & $0.58_{\pm 0.02}$ & $11.31_{\pm 0.12}$ \\
FunSearch + GRPO & $7.59_{\pm 0.34}$ & $6.37_{\pm 0.19}$ & $0.75_{\pm 0.03}$ & $\mathbf{0.51}_{\pm 0.02}$ & $13.41_{\pm 0.15}$ \\
\rowcolor{cityu!8}
FunSearch + ATPO & $\mathbf{6.17}_{\pm 0.96}$ & $\mathbf{6.21}_{\pm 0.28}$ & $0.62_{\pm 0.02}$ & $\mathbf{0.51}_{\pm 0.02}$ & $\mathbf{7.90}_{\pm 0.12}$ \\
\midrule
EoH & $6.33_{\pm 0.40}$ & $6.75_{\pm 0.27}$ & $0.91_{\pm 0.05}$ & $0.84_{\pm 0.13}$ & $12.93_{\pm 0.19}$ \\
EoH + GRPO & $6.31_{\pm 0.44}$ & $7.35_{\pm 0.23}$ & $0.88_{\pm 0.13}$ & $0.79_{\pm 0.13}$ & $\mathbf{8.89}_{\pm 0.96}$ \\
\rowcolor{cityu!8}
EoH + ATPO & $\mathbf{6.29}_{\pm 0.16}$ & $\mathbf{6.51}_{\pm 0.26}$ & $\mathbf{0.69}_{\pm 0.04}$ & $\mathbf{0.73}_{\pm 0.08}$ & $9.05_{\pm 2.00}$ \\
\midrule
OpenEvolve & $6.79_{\pm 0.16}$ & $7.19_{\pm 0.16}$ & $0.79_{\pm 0.02}$ & $0.73_{\pm 0.08}$ & $13.23_{\pm 0.03}$ \\
OpenEvolve + GRPO & $7.25_{\pm 0.21}$ & $6.39_{\pm 0.35}$ & $0.56_{\pm 0.02}$ & $0.59_{\pm 0.02}$ & $14.52_{\pm 0.03}$ \\
\rowcolor{cityu!8}
OpenEvolve + ATPO & $\mathbf{6.27}_{\pm 0.45}$ & $\mathbf{5.97}_{\pm 0.58}$ & $\mathbf{0.54}_{\pm 0.02}$ & $\mathbf{0.55}_{\pm 0.02}$ & $\mathbf{8.35}_{\pm 0.14}$ \\
\bottomrule
\end{tabular}
\endgroup
\label{tab:main_1p5_tsp}
\end{table}

\subsection{Search Performance with ATPO}

To address \textbf{RQ1}, we compare three configurations of each search method on ASP, OBP, and TSP: the baseline method alone, the method combined with standard GRPO, and the method combined with ATPO. We only use Qwen2.5-7B on TSP, as Qwen2.5-1.5B does not reliably generate high-quality algorithms in our tested settings. 
The total evaluation budget for these three problems is 20,000.
We perform three independent runs for each method to account for experimental biases.

\paragraph{Results.}
Table~\ref{tab:main_1p5_tsp} reports results for Qwen2.5-1.5B and Qwen2.5-7B on ASP and OBP, and for Qwen2.5-7B on TSP. We make four observations:
\begin{itemize}[leftmargin=*]
    \item Compared with the corresponding base search methods, ATPO reduces the average relative gap in most settings, with particularly large reductions on TSP. An exception is FunSearch with Qwen2.5-1.5B on OBP, where ATPO performs slightly worse than the base method.
    \item Compared with standard GRPO under the matched evaluation budget, ATPO achieves a lower average relative gap in most settings across the three tasks. It is slightly worse than GRPO with EoH on TSP, while FunSearch with Qwen2.5-7B on OBP yields the same gap at the reported precision. We discuss these exceptions and possible explanations in Appendix~\ref{app:backbone-dependent-results}.
    \item Reinforcement learning does not always improve search performance, as GRPO underperforms the corresponding base search methods in several settings. One possible explanation is that feedback from individual implementations may misrepresent idea quality and steer subsequent search toward less productive directions.
    \item Both Qwen2.5-1.5B and Qwen2.5-7B achieve comparable performance on ASP and OBP. Despite having fewer parameters, Qwen2.5-1.5B can outperform Qwen2.5-7B in some settings. This suggests that larger model size does not necessarily lead to better search performance. Similar observations have also been reported in prior work~\cite{liu2024evolution, zhang2024understanding}.
\end{itemize}

These results suggest that ATPO can improve search efficiency. A possible reason is that the training signal encourages the model to explore more valuable ideas within the current search context, which in turn guides the search toward more promising directions and helps discover better algorithms within the same evaluation budget.

\begin{figure}[t]
    \centering
    \includegraphics[width=0.98\linewidth]{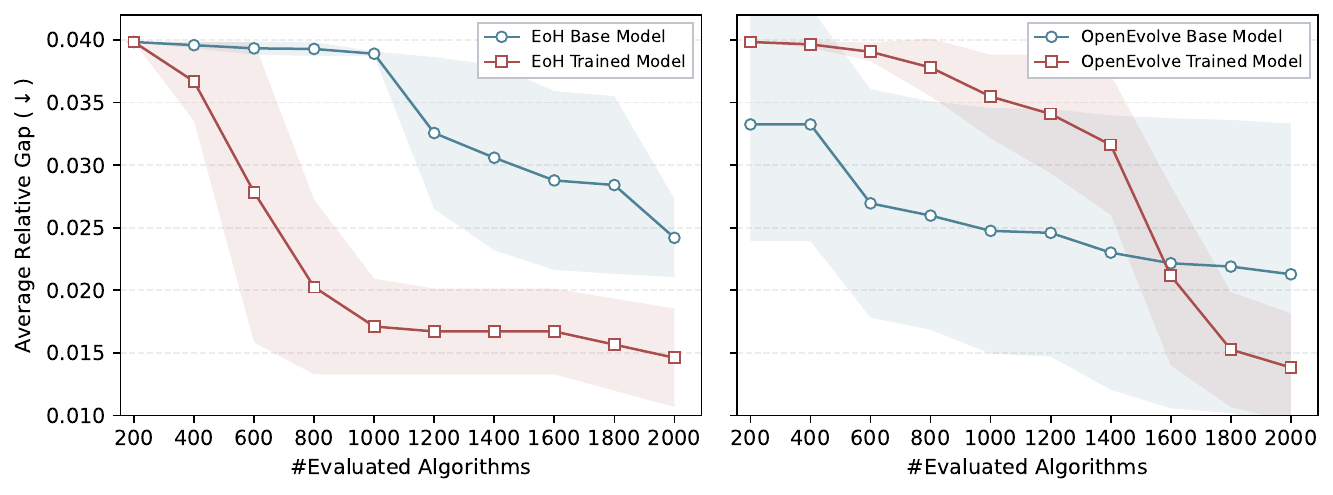}
    \caption{Convergence curve comparison on the performance of top-10 algorithms obtained by EoH and OpenEvolve with different LLM settings. 
    The y-axis reports the average relative gap on the held-out OBP instances, with lower being better. 
    The curve indicates the mean average relative gap over three independent runs, while the shaded regions indicate the standard deviation.}
    \label{fig:obp_transfer}
\end{figure}

\begin{table}[htbp]
\centering
\caption{Average relative gap of the top-5 algorithms on held-out TSP instances using Qwen2.5-7B (\%, lower is better). Results are reported as mean $\pm$ standard deviation over three independent runs. The best mean within each search method is in \textbf{bold}.}
\begin{tabular}{lccc}
\toprule
Model & FunSearch & EoH & OpenEvolve \\
\midrule
Base model & $10.21_{\pm 0.79}$ & $15.20_{\pm 0.18}$ & $14.84_{\pm 0.06}$ \\
GRPO-trained model & $16.11_{\pm 0.04}$ & $13.31_{\pm 0.07}$ & $15.94_{\pm 0.15}$ \\
\rowcolor{cityu!8}
ATPO-trained model & $\mathbf{9.82}_{\pm 0.85}$ & $\mathbf{11.26}_{\pm 2.84}$ & $\mathbf{13.02}_{\pm 0.09}$ \\
\bottomrule
\end{tabular}
\label{tab:topk_tsp_seed2026_grpo}
\end{table}

\subsection{Generalization Across Search Methods and Problem Instances}

To address \textbf{RQ2}, we evaluate whether a model trained with ATPO generalizes to different search methods and to held-out problem instances.
Specifically, the GRPO-trained and ATPO-trained models are the Qwen2.5-7B models obtained in RQ1 under the FunSearch+GRPO and FunSearch+ATPO settings, respectively. We deploy these trained models directly on TSP without any further fine-tuning, using all three search methods. 
Each run uses an evaluation budget of 2,000. 
Table~\ref{tab:topk_tsp_seed2026_grpo} reports the top-5 scores on held-out problem instances generated from a different random seed, covering all three search methods paired with the base model, the GRPO-trained model, and the ATPO-trained model.

We find that the ATPO-trained model consistently outperforms both the base model and the GRPO-trained model on these held-out TSP problem instances across all three search methods. 
This suggests that ATPO improves the generalization of algorithmic reasoning by assigning more reliable credit to promising ideas, thereby reducing the tendency to over-credit implementation-specific behaviors that happen to work on a particular set of sampled problem instances.

Figure~\ref{fig:obp_transfer} further illustrates this transferability on OBP. Here, we use Qwen2.5-1.5B trained with FunSearch+ATPO and deploy it with EoH and OpenEvolve. We test these methods on OBP instances sampled from a different random seed. In both cases, the ATPO-trained model converges faster and reaches a lower average relative gap than the corresponding base model, suggesting that the learned policy transfers both across search methods on the same task and to different problem instances.

Taken together, these results show that a model trained with ATPO on one search method can generalize to different search methods and to unseen problem instances of the same task.

\begin{figure}[t]
    \centering
    \subcaptionbox{Mean reward on ASP over training steps.\label{fig:ablation_asp}}{%
        \includegraphics[width=0.49\linewidth]{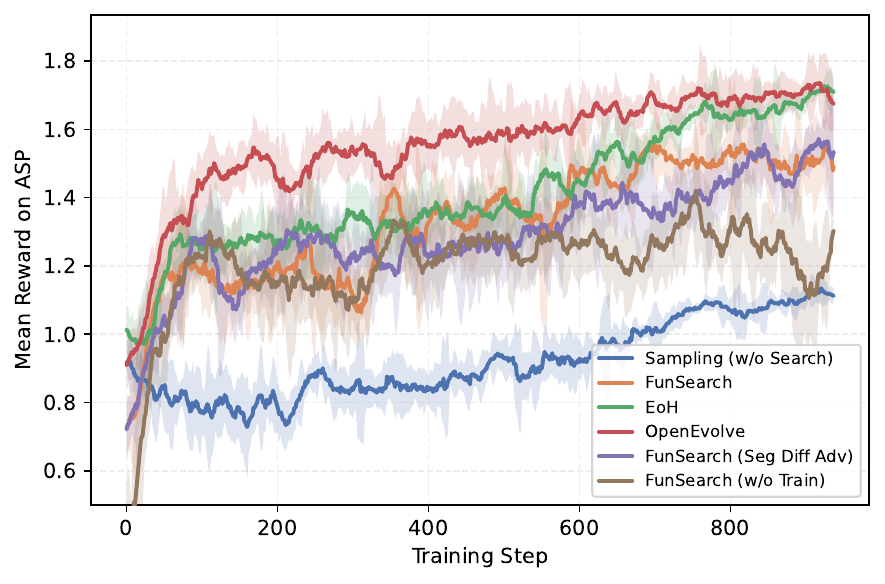}%
    }
    \hfill
    \subcaptionbox{Mean reward on OBP over training steps.\label{fig:ablation_obp}}{%
        \includegraphics[width=0.49\linewidth]{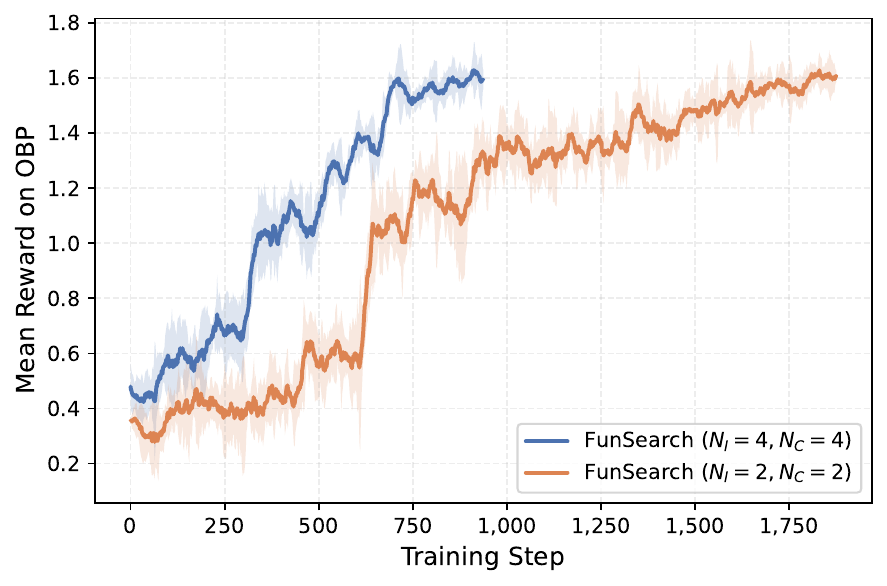}%
    }
    \caption{Ablation study results. For both figures, the y-axis reports the mean of idea- and code-level scores averaged over the rollout batch.
    \textbf{(a)} Mean reward on ASP during training across different search methods and credit-assignment variants. \textbf{(b)} Mean reward on OBP during training comparing two tree size configurations ($N_I \times N_C$).}
    \label{fig:ablation}
\end{figure}

\subsection{Ablation Study}
\label{sec:ablation}

\paragraph{Effect of Search and Training.}
We study the contributions of search and online policy training on ASP using Qwen2.5-1.5B. We evaluate six settings. Three integrate ATPO with FunSearch, EoH, and OpenEvolve, respectively. The remaining settings use policy training through sampling without search, FunSearch without policy updates, and FunSearch with the segment-differentiated advantage variant. Figure~\ref{fig:ablation}(a) reports the reward curves for these settings, from which we draw three observations.

\begin{itemize}[leftmargin=*]
    \item First, removing search (\textit{Sampling (w/o search)}) substantially slows reward improvement. 
    As the population evolves, search framework constructs prompts from different candidate programs, potentially increasing the diversity of the resulting rollout data. 
    Without search, the model trains on a fixed prompt, which may limit rollout diversity and contribute to slower reward improvement.

    \item Second, search without policy updates (\textit{FunSearch (w/o train)}) yields initial reward gains but plateaus later. With ATPO, rewards continue to improve beyond this plateau, indicating that online policy updates help the generator learn more effectively from the collected rollouts.

    \item Third, the two credit-assignment variants introduced in Section~\ref{sec:method} produce reward curves that converge to comparable levels. Specifically, weighted-broadcast advantage (\textit{FunSearch}) and segment-differentiated advantage (\textit{FunSearch (Seg Diff Adv)}) lead to similar final performance. This suggests that both strategies can support effective policy learning in this setting.
\end{itemize}

These results suggest that search and policy training play complementary roles: search supplies evolving algorithmic contexts, while policy updates help the generator learn from the resulting evaluator feedback.

\paragraph{Sensitivity to Tree Size.}
To examine how the numbers of sampled ideas and implementations per prompt affect learning, we compare two tree configurations on OBP: $N_I{=}4, N_C{=}4$ with a batch size of 2, and $N_I{=}2, N_C{=}2$ with a batch size of 4. Figure~\ref{fig:ablation}(b) reports the corresponding reward curves. The larger tree evaluates 32 programs per update step, whereas the smaller tree evaluates 16. As a result, the smaller tree requires roughly twice as many training steps to consume a comparable evaluation budget, which is why the orange curve extends to about twice the number of steps shown for the blue curve. Despite this difference, both configurations converge to similar final reward levels. The larger tree reaches that level in fewer steps, suggesting that greater within-prompt diversity, with more ideas and more implementations per idea, improves convergence efficiency.

\begin{table}[ht]
    \centering
    \caption{Aggregation ablation on ASP with FunSearch + ATPO and Qwen2.5-1.5B. Values are mean $\pm$ standard deviation of the relative gap (\%, lower is better). The best mean is in \textbf{bold}.}
    \begin{tabular}{lc}
        \toprule
        Aggregation method & Average relative gap (\%) \\
        \midrule
        Mean & $6.74_{\pm 0.21}$ \\
        Top-$K$ & $\mathbf{6.17}_{\pm 0.96}$ \\
        \bottomrule
    \end{tabular}
    \label{tab:aggregation_ablation}
\end{table}

\paragraph{Effect of Idea-Score Aggregation.}
Our default top-$K$ scoring estimates an idea's potential through a softmax-weighted aggregation of its top-$K$ code scores. To assess the benefit of emphasizing stronger implementations, we consider mean aggregation as a natural alternative, assigning equal weight to all child-code scores and using their arithmetic mean as the idea score. We conduct this ablation using FunSearch + ATPO on ASP with Qwen2.5-1.5B. As shown in Table~\ref{tab:aggregation_ablation}, top-$K$ achieves a lower mean relative gap, suggesting that emphasizing stronger implementations helps estimate an idea's achievable implementation potential.

Due to space limitations, additional experiments and analysis are provided in Appendix~\ref{app:additional-experiments}.

%% file: src/5-conclusion.tex
\section{Conclusion}
\vspace{-0.5em}
In this paper, we introduce ATPO, a reinforcement learning method that can be integrated into the algorithm-generation stage of existing LLM-AAD search frameworks. ATPO uses tree-based sampling to explore multiple ideas under each prompt and multiple code implementations under each idea. It aggregates evaluator feedback to estimate each idea's achievable implementation potential and assigns credit through inter-idea and within-idea comparisons, enabling the LLM to learn from search outcomes and better guide subsequent algorithm generation. Experiments with FunSearch, EoH, and OpenEvolve on ASP, OBP, and TSP show improved search performance in most settings under a fixed evaluation budget. The learned policy also transfers to held-out instances and alternative search methods within the same task.

\paragraph{Limitations.}
ATPO estimates an idea's achievable implementation potential from a finite set of sampled implementations. 
Although multiple implementations improve score stability, the estimate remains sensitive to implementation quality and sampling randomness. 
Marker-based parsing does not guarantee clean semantic separation, as models may place idea or implementation content in the wrong segment.
Empirically, ATPO does not outperform the base search or GRPO in every setting. Our comparisons control the evaluator-call budget, but do not establish an advantage under equal wall-clock time, generated-token budgets, or GPU-hours. 
Finally, the transfer evidence is limited to held-out instances and alternative search frameworks within the same task. 

\paragraph{Future Work.}
We identify two main directions for future work.
\ding{182}~Adaptively allocate implementation samples according to the uncertainty of idea-level estimates, balancing estimation reliability against evaluation cost.
\ding{183}~Better integrate ATPO with existing LLM-AAD search methods by coordinating policy updates with the search process. Since these methods were not originally designed for online policy training, improving the cleaning and deduplication of rollout data is a practical step toward this integration. 

%% file: src/6-appendix.tex
\section{Implementation Details}
\label{app:implementation}

\subsection{Tree Generation} 
\label{app:tree-gen}

\begin{figure}[t]
    \centering
    \includegraphics[width=\linewidth]{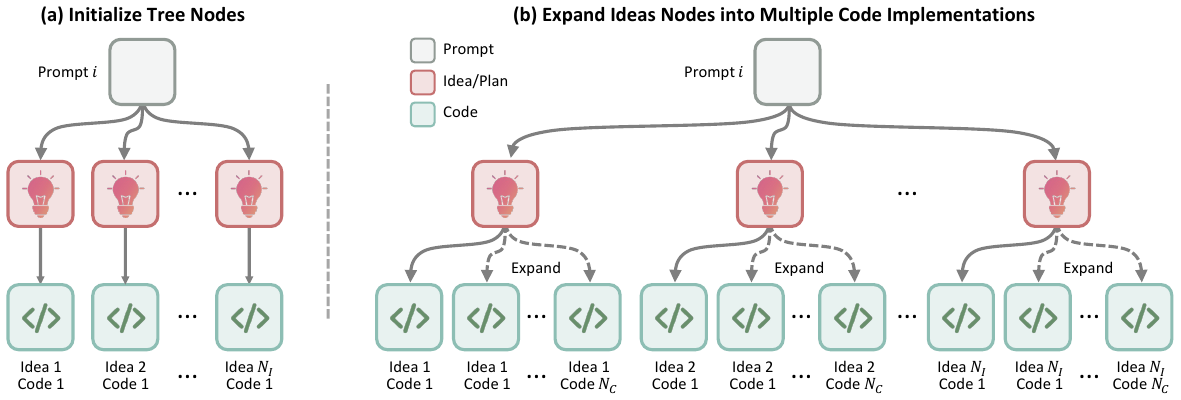}
    \caption{Two-phase tree construction. \textbf{(a)}~Initialization: for each prompt, the LLM first generates $N_I$ idea nodes, each paired with one initial code candidate. \textbf{(b)}~Expansion: each valid idea node is expanded with $N_C - 1$ additional code candidates conditioned on the same idea, completing the full $N_I \times N_C$ leaf tree.}
    \label{fig:tree_gen}
\end{figure}

Figure~\ref{fig:tree_gen} illustrates the two-phase construction used to build the idea--code tree for each prompt.
To enable separate extraction of idea and code nodes, we modify the search method's prompts to explicitly instruct the model to place its idea and implementation under the \texttt{\#\#\# Idea} and \texttt{\#\#\# Code} headings, respectively.

\paragraph{Phase (a): initialization.}
The model generates $N_I$ responses for prompt $p_i$ in a single batched forward pass, each containing both an idea plan and an initial code candidate.
Each response is split at the \texttt{\#\#\# Code} delimiter: tokens before it form the idea segment and tokens from \texttt{\#\#\# Code} onward form the code segment.
This produces a skeleton tree of $N_I$ idea nodes, each with one child code node (Idea~1 Code~1, \ldots, Idea~$N_I$ Code~1 in panel~(a)).
Responses missing either delimiter are treated as invalid leaves, assigned a fixed negative fallback score, and not expanded in Phase~(b). They are still retained in the batch to provide direct supervision on format failures.

\paragraph{Phase (b): expansion.}
For each valid idea node, the remaining $N_C - 1$ code candidates are generated by conditioning the model on the concatenation of the original prompt and the idea text.
This keeps the idea fixed while sampling diverse implementations, expanding each idea into $N_C$ code children and completing the full $N_I \times N_C$ leaf tree shown in panel~(b).

\subsubsection{Handling Format Errors}
\label{app:format-errors}

We explicitly detect malformed generations and terminate them early instead of trying to repair them heuristically.
During Phase~(a), a response is considered valid only if it contains both the \texttt{\#\#\# Idea} and \texttt{\#\#\# Code} delimiters.
If either delimiter is missing, we convert the entire response into a single depth-1 leaf, mark it as an invalid-format sample, and exclude it from the Phase~(b) expansion step.
In other words, malformed initialization trajectories do not produce additional code children.

During Phase~(b), the idea text is already fixed by a valid parent node, so we only check whether the newly generated continuation contains the \texttt{\#\#\# Code} delimiter.
If this delimiter is missing, we still keep the resulting leaf in the rollout batch, but we record an empty code field and mark the sample as an invalid format.
These invalid leaves are retained so that the policy still receives direct supervision, as such outputs are undesirable.

In the reward pipeline, invalid-format leaves are assigned a dedicated state before program extraction and execution.
They bypass both code extraction and downstream evaluation and receive a fixed fallback score of $-1.0$.
This design has two effects: it prevents malformed generations from consuming evaluator budget, and it turns formatting mistakes into explicit negative training signals rather than silently dropping them from the batch.

\begin{figure}[t]
    \centering
    \includegraphics[width=\linewidth]{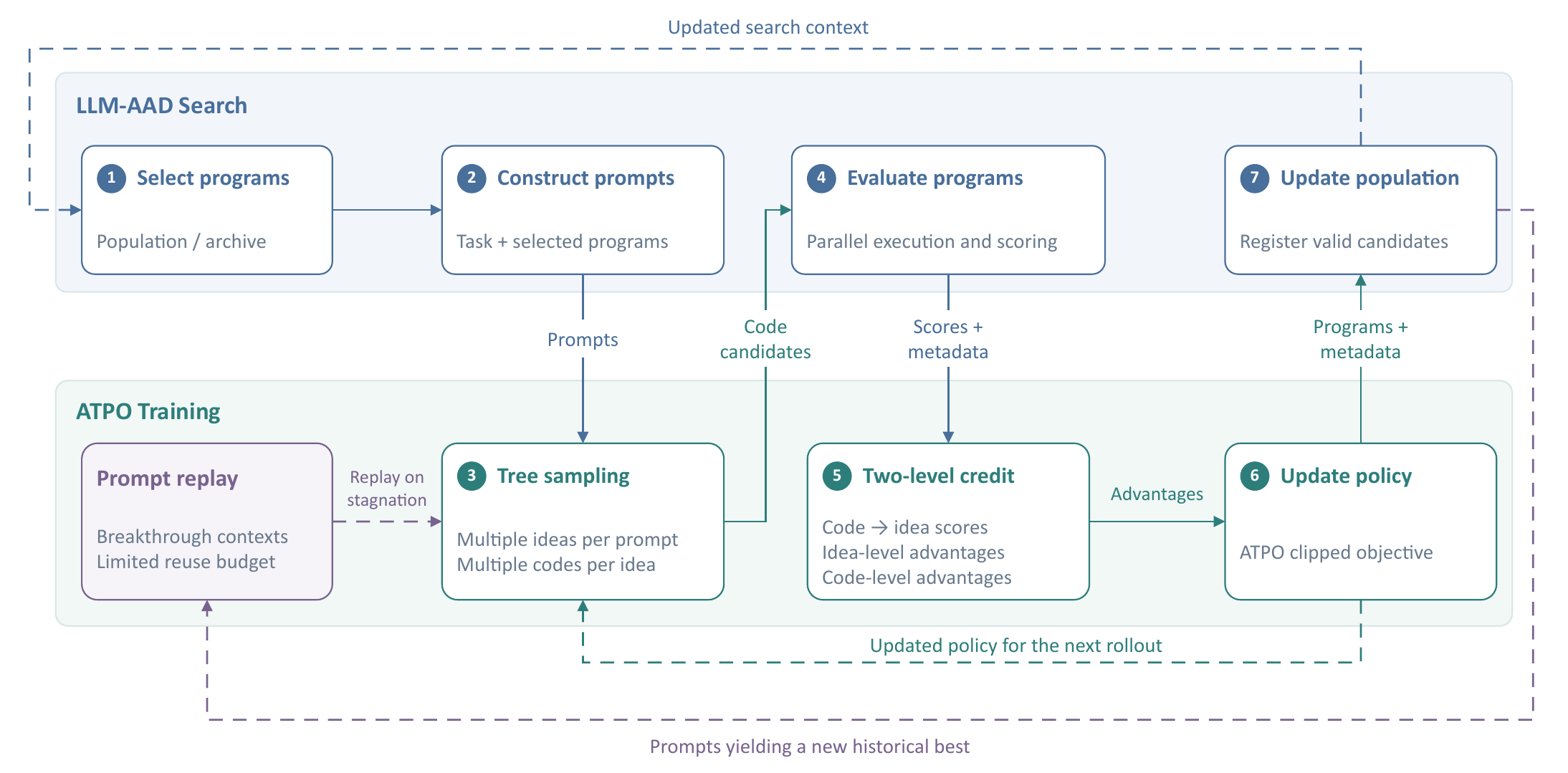}
    \caption{Workflow of ATPO integrated with an LLM-AAD search method. The upper row shows the LLM-AAD search process, including program selection, prompt construction, evaluation, and population update. The lower row shows tree sampling, two-level credit assignment, policy update, and prompt replay. Solid arrows indicate the main data flow between these stages, while dashed arrows indicate feedback through the updated search context and policy, as well as the storage and replay of breakthrough prompts.}
    \label{fig:algo_tree_rl_pipeline}
\end{figure}

\subsection{Replay Buffer Details}
\label{app:replay-buffer}

The data-generation process of LLM-AAD search methods is driven by the evolving population, and the prompt contexts supplied to the policy change continuously as search progresses.
As a result, the prompt distribution is non-stationary and stochastic: whether a breakthrough prompt is revisited in the future is not guaranteed by the search method alone.
The replay buffer addresses this by explicitly retaining prompt contexts that led to genuine historical progress, allowing the controller to revisit them when the search stagnates.
This mechanism encourages the policy to re-explore previously fruitful directions rather than drifting away from them as the search population evolves.

ATPO-AAD pipeline maintains a replay buffer of \emph{breakthrough prompts}, i.e., prompt contexts that lead to genuine historical progress during online search. Let $s_t^{\max}$ denote the best code score produced at step $t$, and let $b_{t-1}$ be the best score observed before that step. If $s_t^{\max} > b_{t-1}$, we insert the corresponding prompt into the buffer together with its score and a usage counter. The buffer, therefore, stores only prompt contexts that improved the search frontier, rather than all prompts or all high-scoring prompts.

Replay is activated only when the online search stagnates. Let $c_t$ be the number of consecutive iterations without improving the historical best score. Once $c_t$ exceeds a preset threshold, the controller switches to replay mode with probability $\rho_{\mathrm{replay}}$ and replaces the standard prompt-selection step with a prompt sampled from the replay buffer. This mechanism revisits prompt contexts that drive breakthroughs and helps the search escape from locally stable but mediocre regions.

Each replay prompt has a finite lifetime. Every entry maintains a usage count $u$ and remains active only while $u < U_{\max}$. Whenever a prompt is replayed, its usage count is incremented. Once the count reaches $U_{\max}$, the prompt is permanently retired. In all experiments, we use the default setting $U_{\max}=1$, meaning that each stored breakthrough prompt is replayed at most once. This tight budget prevents the search from over-committing to a small number of lucky prompts while still preserving useful historical context.

\subsection{Integration with LLM-AAD Search}
\label{app:system-workflow}

Figure~\ref{fig:algo_tree_rl_pipeline} summarizes how ATPO is integrated into an LLM-AAD search loop. In the upper row, the search method selects programs from the population, constructs prompts, evaluates generated code candidates in parallel, and registers valid candidates to update the search context. In the lower row, the training framework expands each prompt into multiple ideas and code implementations, aggregates code scores into idea scores, computes idea-level and code-level advantages, and updates the policy with the ATPO clipped objective.

The solid arrows trace the exchange of prompts, code candidates, evaluation scores, and program metadata between the two components. The dashed feedback paths carry the updated search context and policy into subsequent iterations. Prompts that yield a new historical best are stored in the replay buffer and revisited when search stagnates, subject to a limited reuse budget. This interface preserves the search method's role as the outer exploration controller while allowing the policy to learn from trajectories collected during search.

\section{Additional Experiments}
\label{app:additional-experiments}

\subsection{Additional Ablation Results}
\label{app:ablation-replay}

\paragraph{Replay Buffer Ablation.}

\begin{figure}[t]
    \centering
    \includegraphics[width=0.6\linewidth]{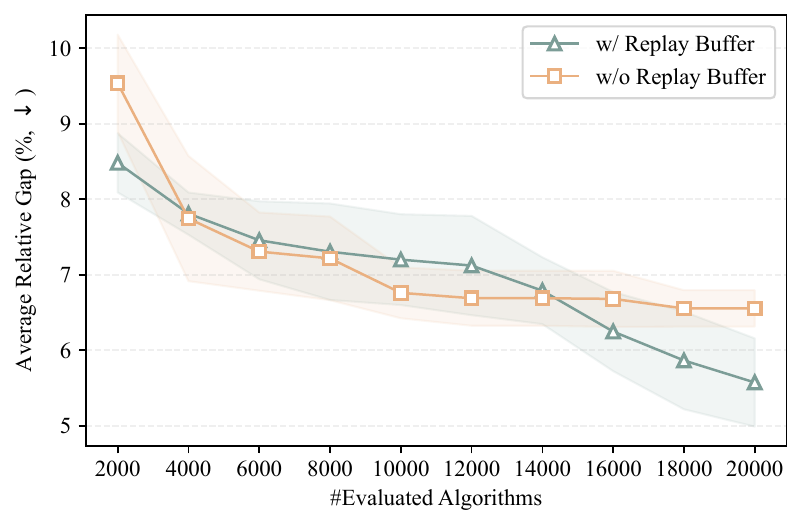}
    \caption{Top-30 average relative gap convergence curves on ASP comparing ATPO with and without the replay buffer. The experiment is conducted with FunSearch + ATPO using Qwen2.5-1.5B as the base model. The x-axis reports the number of evaluated algorithms, and the y-axis reports the average relative gap (\%) of the top-30 programs discovered so far, where lower is better.}
    \label{fig:ablation_replay}
\end{figure}

Figure~\ref{fig:ablation_replay} compares ATPO with and without the replay buffer on ASP, measured by the average relative gap of the top-30 programs over the course of search.
Both variants improve steadily in the early stage, but the version without the replay buffer plateaus earlier and at a higher gap.
The variant with the replay buffer continues to improve in the later stages, suggesting that revisiting breakthrough prompt contexts helps the policy escape stagnation and exploit historically productive directions more effectively.

\subsection{Variation Across Search Backbones and Tasks}
\label{app:backbone-dependent-results}

As shown in Table~\ref{tab:main_1p5_tsp}, ATPO improves performance in most evaluated settings, but does not uniformly outperform either the base search method or standard GRPO. On OBP, FunSearch with Qwen2.5-1.5B obtains a gap of $0.57\%$, compared with $0.62\%$ for FunSearch + ATPO. The gaps on this benchmark are already small, leaving limited room for improvement. Stochastic sampling and search may therefore contribute to close or occasionally negative results, although proximity to the reference bound alone does not establish the cause of this exception.

On TSP, EoH + ATPO obtains a gap of $9.05\%$, slightly above the $8.89\%$ of EoH + GRPO in Table~\ref{tab:main_1p5_tsp}, whereas ATPO outperforms GRPO with FunSearch and OpenEvolve. To examine this difference, Table~\ref{tab:backbone_training_diagnostics} reports training diagnostics over the final quarter of training. These diagnostics use the original reward scale for top-5 performance, rather than the relative gaps in Table~\ref{tab:main_1p5_tsp}.

\begin{table}[htbp]
    \centering
    \caption{TSP training diagnostics across search backbones over the final quarter of training. Top-5 values are rewards on the original evaluator scale (higher is better). ``Adv.'' denotes mean absolute combined advantage. The idea/code column reports the ratio of idea-level to code-level advantage norms.}
    \label{tab:backbone_training_diagnostics}
    \begingroup
    \small
    \setlength{\tabcolsep}{4pt}
    \begin{tabular}{lcccccc}
        \toprule
        Backbone & \makecell{ATPO\\top-5} & \makecell{GRPO\\top-5} & $\lVert A_{\mathrm{idea}}\rVert/\lVert A_{\mathrm{code}}\rVert$ & \makecell{ATPO\\adv.} & \makecell{GRPO\\adv.} & \makecell{GRPO\\spread} \\
        \midrule
        FunSearch & 0.671 & 0.565 & 2.05 & 0.017 & 0.012 & 0.39 \\
        EoH & 0.631 & 0.652 & 3.43 & 0.076 & 0.091 & 2.62 \\
        OpenEvolve & 0.663 & 0.543 & 2.81 & 0.104 & 0.033 & 1.79 \\
        \bottomrule
    \end{tabular}
    \endgroup
\end{table}

As shown in Table~\ref{tab:backbone_training_diagnostics}, EoH exhibits a larger mean absolute GRPO advantage and a greater GRPO reward spread than the other two search backbones, supporting the observation of a stronger and more dispersed outcome-level signal. One possible explanation is that EoH's diverse evolutionary operators already provide informative variation in generated ideas, reducing the marginal benefit of hierarchical credit assignment. This is a post hoc interpretation supported by the diagnostics, rather than a causal conclusion. More generally, the benefit of ATPO depends on the feedback induced by the search backbone, and online policy updates do not guarantee improvement in every setting.

\subsection{Stability of Idea-Level Scores}
\label{app:idea-score-stability}

We examine whether using multiple implementations makes idea-level scores more stable under independent code resampling. For each of 20 prompts, we first generate candidate ideas and then hold their text fixed across five independent implementation-resampling rounds. In each round, we sample code conditioned on the same prompt and idea, using 1, 4, or 8 implementations per idea.

We assess stability using top-2 overlap, which measures agreement between the two highest-ranked ideas across rounds, and mean absolute score difference, which measures changes in the score assigned to the same idea. Higher overlap and lower score difference indicate greater stability. These are diagnostics of idea-score estimation on its original score scale, rather than task-level relative gaps.

\begin{table}[ht]
    \centering
    \caption{Stability of idea-level scores under independent implementation resampling. The single-implementation setting corresponds to scoring each idea from one outcome, as in linear GRPO.}
    \begin{tabular}{ccc}
        \toprule
        Implementations per idea & Top-2 overlap ($\uparrow$) & Mean absolute score difference ($\downarrow$) \\
        \midrule
        1 (GRPO) & 0.574 & 0.659 \\
        4 & 0.613 & 0.176 \\
        8 & 0.655 & 0.047 \\
        \bottomrule
    \end{tabular}
    \label{tab:idea_score_stability}
\end{table}

As shown in Table~\ref{tab:idea_score_stability}, increasing the number of implementations improves ranking agreement and reduces score variation across resampling rounds. This supports the use of multiple implementations for more stable estimation, while leaving uncertainty about an idea's potential under finite sampling. It does not establish an implementation-independent ground-truth quality for an idea.

We also compare top-$K$ and mean aggregation with eight implementations per idea (Table~\ref{tab:aggregation_stability}). Top-$K$ produces slightly higher ranking agreement and slightly lower score variation. Together with the downstream comparison in Table~\ref{tab:aggregation_ablation}, these results support top-$K$ as a practical scoring choice, while the stability differences between the two aggregators remain modest.

\begin{table}[ht]
    \centering
    \caption{Stability comparison of top-$K$ and mean aggregation with eight implementations per idea. The best value in each column is in bold.}
    \begin{tabular}{lcc}
        \toprule
        Aggregation & Top-2 overlap $\uparrow$ & Mean absolute score difference $\downarrow$ \\
        \midrule
        Mean & 0.640 & 0.051 \\
        Top-$K$ & $\mathbf{0.655}$ & $\mathbf{0.047}$ \\
        \bottomrule
    \end{tabular}
    \label{tab:aggregation_stability}
\end{table}

\section{Task Details, Prompt Templates, and Reward Formulation}
\label{app:task-prompts}

For completeness, we collect in one place the exact task-facing descriptions, bundled template programs, and reward formulations for the three evaluation tasks in Section~\ref{sec:experiments}. We present the prompt descriptions and templates in compact boxed form because these assets define the program interface exposed to the search policy.

\lstdefinestyle{tasklisting}{
    basicstyle=\ttfamily\footnotesize,
    breaklines=true,
    columns=fullflexible,
    keepspaces=true,
    upquote=true,
    showstringspaces=false
}

\newtcblisting{taskbox}[2][]{%
    breakable,
    listing only,
    listing engine=listings,
    colback=cityu!4,
    colframe=cityu!75!black,
    boxrule=0.5pt,
    arc=1mm,
    left=1mm,
    right=1mm,
    top=0.5mm,
    bottom=0.5mm,
    title={#2},
    fonttitle=\bfseries,
    coltitle=white,
    listing options={style=tasklisting},
    #1
}

\subsection{Admissible Set}

This task asks the model to implement a scalar priority function for vectors in $\{0,1,2\}^{15}$ with exactly ten non-zero entries. The evaluator greedily builds an admissible set by ranking candidate vectors with the generated priority and filtering out those that violate the combinatorial constraints.

\begin{taskbox}{Admissible Set: Task Description}
Formally, admissible set problems, denoted as A(n, w), are collections of
vectors in {0, 1, 2}^n that satisfy the following constraints:

1. Each vector is unique and has exactly w non-zero coordinates.
2. For any three distinct vectors, there exists a coordinate whose three values
   are one of the following patterns: {0, 1, 2}, {0, 0, 1}, or {0, 0, 2}.

The objective is to maximize the size of the admissible set while satisfying
all constraints. In this task, use n = 15 and w = 10.

You must implement a function named "priority". Do not change this function
name, because the evaluator will extract and call "priority" directly.

Recommended signature:
- "def priority(el: tuple[int, ...], n: int = 15, w: int = 10) -> float"

Your job is to design this "priority" function so it helps construct a large
admissible set.
\end{taskbox}

\begin{taskbox}{Admissible Set: Template Program}
import math
import numpy as np

# Do not rename "priority". The evaluator extracts this exact function name.
def priority(el: tuple[int, ...], n: int = 15, w: int = 10) -> float:
    """Return the priority used to rank a candidate vector.

    Args:
        el: Candidate vector represented as a tuple in {0, 1, 2}^n.
        n: Length of the vector.
        w: Number of non-zero coordinates in the vector.

    Returns:
        A scalar score. Larger values mean the candidate is selected earlier.
    """
    return 0.0
\end{taskbox}

\subsection{Online Bin Packing}

In the online bin packing task, items arrive sequentially and must be assigned immediately to one of the currently feasible bins. The generated program outputs a priority score for each valid bin, and the evaluator places the item into the highest-scoring option.

\begin{taskbox}{Online Bin Packing: Task Description}
Online bin packing receives items one by one and must place each item
immediately, without revisiting earlier decisions.

You must implement a function named "priority". Do not change this function
name, because the evaluator will extract and call "priority" directly.

Recommended signature:
- "def priority(item: float, bins: np.ndarray) -> np.ndarray"

Design this "priority" function so it scores every currently valid bin for the
incoming item. The evaluator assigns the item to the valid bin with the highest
priority score.

Inputs:
- "item": the size of the arriving item.
- "bins": a NumPy array containing the remaining capacities of all bins that can
  still fit the item.

Output:
- A NumPy array of priority scores with the same shape as "bins".

Goal:
- Minimize the total number of bins used across the evaluation dataset.
\end{taskbox}

\begin{taskbox}{Online Bin Packing: Template Program}
import numpy as np

# Do not rename "priority". The evaluator extracts this exact function name.
def priority(item: float, bins: np.ndarray) -> np.ndarray:
    """Return a priority score for each currently valid bin.

    Args:
        item: Size of the arriving item.
        bins: Remaining capacities of the bins that can still fit the item.

    Returns:
        One score per valid bin. Larger values mean the bin is preferred.
    """
    return -(bins - item)
\end{taskbox}

\subsection{Constructive TSP}

For constructive TSP, the model implements a node-selection rule that extends a tour one step at a time from the current city to one unvisited city. The evaluator repeatedly queries this rule on random Euclidean instances and scores the resulting heuristic by the average completed tour length.

\begin{taskbox}{Constructive TSP: Task Description}
Given a set of nodes with 2D coordinates, construct a Traveling Salesman tour
that visits each node exactly once and returns to the starting node.

You must implement a function named "select_next_node". Do not change this
function name, because the evaluator will extract and call "select_next_node"
directly.

Recommended signature:
- "def select_next_node(current_node, destination_node, unvisited_nodes, distance_matrix) -> int"

The evaluator solves the problem step by step:
- It starts from node "0".
- At each step it calls "select_next_node(...)".
- Your function must choose one node from "unvisited_nodes".

Inputs:
- "current_node": index of the current node
- "destination_node": index of the designated destination node
- "unvisited_nodes": NumPy array of currently unvisited node indices
- "distance_matrix": full pairwise distance matrix

Output:
- The index of the next node to visit

Goal:
- Minimize the average tour length across the bundled random instances.
\end{taskbox}

\begin{taskbox}{Constructive TSP: Template Program}
import numpy as np

# Do not rename "select_next_node". The evaluator extracts this exact function name.
def select_next_node(
    current_node: int,
    destination_node: int,
    unvisited_nodes: np.ndarray,
    distance_matrix: np.ndarray,
) -> int:
    """Select the next node in a constructive TSP tour.

    Args:
        current_node: Index of the current node.
        destination_node: Index of the fixed destination node.
        unvisited_nodes: Array of node indices that have not been visited yet.
        distance_matrix: Pairwise Euclidean distance matrix.

    Returns:
        The index of the next node to visit.
    """
    next_node = unvisited_nodes[0]
    return next_node
\end{taskbox}

\subsection{Reward Formulation}
\label{app:reward}

All three tasks share a common reward structure: the task evaluator executes the generated code and returns a normalized scalar score $r \in [0, 1]$, which serves directly as the code-level reward in ATPO.
The normalization anchors each score against a baseline heuristic (the bundled template) and a theoretical performance bound, ensuring that the reward is comparable across tasks and that improvements over the template are meaningfully distinguished.

\paragraph{Admissible Set.}
\label{app:reward-admissible}

The evaluator uses the generated \texttt{priority} function to greedily construct an admissible set $\mathcal{A} \subseteq \{0,1,2\}^{15}$ of weight-10 vectors, where any three vectors must be mutually distinguishable in at least one coordinate.
The reward is the fraction of the known reference optimum achieved:
\begin{equation}
    r = \frac{|\mathcal{A}|}{3003},
\end{equation}
where 3003 is the size of the optimal admissible set for parameters $(n{=}15,\, w{=}10)$.
A score of 1.0 indicates that the greedy construction recovers the full optimal set.

\paragraph{Online Bin Packing.}
\label{app:reward-binpacking}

The evaluator runs the generated \texttt{priority} function on 5 instances of 5{,}000 items drawn from a Weibull distribution (shape 3, capacity 100) and records the mean number of bins used $\bar{b}$.
The reward is defined by a piecewise linear function anchored at three points: the worst-case bound $b_{\text{worst}} = 5000$ (one item per bin), the template score $b_{\text{tpl}}$ (mean bins used by the best-fit heuristic), and the L1 lower bound $b_{\text{best}} = \lceil \sum_i s_i / 100 \rceil$ (mean over instances):
\begin{equation}
    r =
    \begin{cases}
        0 & \text{if } \bar{b} \ge b_{\text{worst}}, \\[4pt]
        0.3 \cdot \dfrac{b_{\text{worst}} - \bar{b}}{b_{\text{worst}} - b_{\text{tpl}}} & \text{if } b_{\text{tpl}} \le \bar{b} < b_{\text{worst}}, \\[8pt]
        0.3 + 0.7 \cdot \dfrac{b_{\text{tpl}} - \bar{b}}{b_{\text{tpl}} - b_{\text{best}}} & \text{if } \bar{b} < b_{\text{tpl}}.
    \end{cases}
\end{equation}
Matching the template yields $r = 0.3$. Reaching the L1 bound yields $r = 1.0$.
The two-segment design places a larger fraction of the reward range (0.7) in the region above the template, providing stronger gradient signal for improvements beyond the baseline.

\paragraph{Constructive TSP.}
\label{app:reward-tsp}

The evaluator runs the generated \texttt{select\_next\_node} function to construct tours on 16 random Euclidean instances of 50 cities and records the mean tour length $\bar{d}$.
The reward follows the same piecewise linear scheme anchored at a worst-case distance $d_{\text{worst}} = 30.0$, the template tour length $d_{\text{tpl}}$, and a best-case anchor $d_{\text{best}} = 5.0$:
\begin{equation}
    r =
    \begin{cases}
        0 & \text{if } \bar{d} \ge d_{\text{worst}}, \\[4pt]
        0.3 \cdot \dfrac{d_{\text{worst}} - \bar{d}}{d_{\text{worst}} - d_{\text{tpl}}} & \text{if } d_{\text{tpl}} \le \bar{d} < d_{\text{worst}}, \\[8pt]
        0.3 + 0.7 \cdot \dfrac{d_{\text{tpl}} - \bar{d}}{d_{\text{tpl}} - d_{\text{best}}} & \text{if } \bar{d} < d_{\text{tpl}}.
    \end{cases}
\end{equation}
Since shorter tours are better, the reward increases as $\bar{d}$ decreases, with $r > 1.0$ possible for heuristics that surpass $d_{\text{best}} = 5.0$.

\section{Use of Large Language Models}
Large Language Models are employed in two ways in this work. (i) We use LLMs to aid or polish writing in this paper. (ii) This work investigates LLM-based automated algorithm design. Therefore, LLMs are used in the experiments.

\section{Broader Impacts}

This work introduces Algorithm Tree Policy Optimization (ATPO), a reinforcement learning method that can be integrated into the algorithm generation stage of existing LLM-based Automated Algorithm Design (LLM-AAD) search frameworks.
By providing segment-level credit assignment at both the idea and code levels, ATPO enables LLMs to learn from evaluator feedback during algorithm search.
This capability may help reduce reliance on human expertise and hand-crafted heuristics in solving complex optimization problems, with demonstrated improvements on benchmarks including the Admissible Set Problem, Online Bin Packing, and the Traveling Salesman Problem.
We hope this method will encourage further research into training LLMs for algorithm design and planning tasks beyond those studied here.
As a method for improving LLM-AAD search frameworks, ATPO is not anticipated to have specific negative societal impacts.

%% file: checklist.tex
\section*{NeurIPS Paper Checklist}

\begin{enumerate}

\item {\bf Claims}
    \item[] Question: Do the main claims made in the abstract and introduction accurately reflect the paper's contributions and scope?
    \item[] Answer: \answerYes{} 
    \item[] Justification: The abstract and introduction accurately outline our contributions and the specific problem settings we address.

\item {\bf Limitations}
    \item[] Question: Does the paper discuss the limitations of the work performed by the authors?
    \item[] Answer: \answerYes{} 
    \item[] Justification: The limitations of this work are discussed in the conclusion section of the paper.

\item {\bf Theory assumptions and proofs}
    \item[] Question: For each theoretical result, does the paper provide the full set of assumptions and a complete (and correct) proof?
    \item[] Answer: \answerNA{} 
    \item[] Justification: The paper does not include theoretical results.

    \item {\bf Experimental result reproducibility}
    \item[] Question: Does the paper fully disclose all the information needed to reproduce the main experimental results of the paper to the extent that it affects the main claims and/or conclusions of the paper (regardless of whether the code and data are provided or not)?
    \item[] Answer: \answerYes{} 
    \item[] Justification: The experiment section and the supplementary material provide details on the method, LLMs used, hyperparameter configurations, and problem specifications.

\item {\bf Open access to data and code}
    \item[] Question: Does the paper provide open access to the data and code, with sufficient instructions to faithfully reproduce the main experimental results, as described in supplemental material?
    \item[] Answer: \answerNo{} 
    \item[] Justification: The experimental tasks and evaluation protocols are established benchmarks adapted from existing public literature, requiring no new dataset generation. We intend to release our full implementation code and configuration details upon the conclusion of the review process. 

\item {\bf Experimental setting/details}
    \item[] Question: Does the paper specify all the training and test details (e.g., data splits, hyperparameters, how they were chosen, type of optimizer) necessary to understand the results?
    \item[] Answer: \answerYes{} 
    \item[] Justification: We have provided the key hyperparameters in the experiment section.

\item {\bf Experiment statistical significance}
    \item[] Question: Does the paper report error bars suitably and correctly defined or other appropriate information about the statistical significance of the experiments?
    \item[] Answer: \answerYes{} 
    \item[] Justification: In our experiments, we report the mean performance and standard deviation aggregated over three runs.

\item {\bf Experiments compute resources}
    \item[] Question: For each experiment, does the paper provide sufficient information on the computer resources (type of compute workers, memory, time of execution) needed to reproduce the experiments?
    \item[] Answer: \answerYes{} 
    \item[] Justification: We have reported the specifications of the GPU server used in this work.
    
\item {\bf Code of ethics}
    \item[] Question: Does the research conducted in the paper conform, in every respect, with the NeurIPS Code of Ethics \url{https://neurips.cc/public/EthicsGuidelines}?
    \item[] Answer: \answerYes{} 
    \item[] Justification: We have reviewed the NeurIPS Code of Ethics and confirm that our research adheres to all prescribed ethical standards and practices.

\item {\bf Broader impacts}
    \item[] Question: Does the paper discuss both potential positive societal impacts and negative societal impacts of the work performed?
    \item[] Answer: \answerYes{} 
    \item[] Justification: We have discussed the broader impacts in the appendix.
    
\item {\bf Safeguards}
    \item[] Question: Does the paper describe safeguards that have been put in place for responsible release of data or models that have a high risk for misuse (e.g., pre-trained language models, image generators, or scraped datasets)?
    \item[] Answer: \answerNA{} 
    \item[] Justification: The paper poses no such risks.

\item {\bf Licenses for existing assets}
    \item[] Question: Are the creators or original owners of assets (e.g., code, data, models), used in the paper, properly credited and are the license and terms of use explicitly mentioned and properly respected?
    \item[] Answer: \answerYes{} 
    \item[] Justification: We use the Qwen model in our project. We follow the Qwen License and have added corresponding citations.

\item {\bf New assets}
    \item[] Question: Are new assets introduced in the paper well documented and is the documentation provided alongside the assets?
    \item[] Answer: \answerNA{} 
    \item[] Justification: This paper does not release new standalone datasets or code in the current submission. All necessary implementation details are fully described in the paper, and the source code will be made publicly available upon the conclusion of the review process.

\item {\bf Crowdsourcing and research with human subjects}
    \item[] Question: For crowdsourcing experiments and research with human subjects, does the paper include the full text of instructions given to participants and screenshots, if applicable, as well as details about compensation (if any)? 
    \item[] Answer: \answerNA{} 
    \item[] Justification: The paper does not involve crowdsourcing nor research with human subjects

\item {\bf Institutional review board (IRB) approvals or equivalent for research with human subjects}
    \item[] Question: Does the paper describe potential risks incurred by study participants, whether such risks were disclosed to the subjects, and whether Institutional Review Board (IRB) approvals (or an equivalent approval/review based on the requirements of your country or institution) were obtained?
    \item[] Answer: \answerNA{} 
    \item[] Justification: This research is purely computational and does not involve any human subjects or crowdsourcing experiments.

\item {\bf Declaration of LLM usage}
    \item[] Question: Does the paper describe the usage of LLMs if it is an important, original, or non-standard component of the core methods in this research? Note that if the LLM is used only for writing, editing, or formatting purposes and does \emph{not} impact the core methodology, scientific rigor, or originality of the research, declaration is not required.
    \item[] Answer: \answerYes{} 
    \item[] Justification: We have provided a section in the Appendix to describe the usage of LLMs.

\end{enumerate}